%% file: main.tex
\documentclass[VANCOUVER,Times2COL]{WileyNJDv5}

\articletype{Article Type}%

\received{Date Month Year}
\revised{Date Month Year}
\accepted{Date Month Year}
\journal{Journal}
\volume{00}
\copyyear{2025}
\startpage{1}

\usepackage{xurl}             
\usepackage{caption}
\usepackage{float}

\usepackage{cuted} 

\usepackage{subscript} 

\usepackage{comment} 

\begin{document}
\hypersetup{hidelinks}

\title{Credible CO2 Comparisons: A Machine Learning Approach to Vehicle Powertrain Assessment}

\author[1]{Rodrigo Pereira David}

\author[1,2]{Luciano Araujo Dourado Filho}

\author[1]{Daniel Marques da Silva}

\author[4]{João Alfredo Cal-Braz}

\authormark{TAYLOR \textsc{et al.}}
\titlemark{PLEASE INSERT YOUR ARTICLE TITLE HERE}

\address[1]{ \orgname{National Institute of Metrology, Quality and Technology}, \orgaddress{\state{Av. N. Sra. das Graças, 50, Duque de Caxias, Rio de Janeiro}, \country{Brazil}}}

\address[2]{\orgname{University of Feira de Santana}, \orgaddress{\state{Av. Transnordestina S/N, Feira de Santana, Bahia}, \country{Brazil}}}

\corres{Corresponding author Rodrigo Pereira David,  \email{rdavidrd@gmail.com}}



\abstract[Abstract]{
Decarbonizing road transport requires consistent and transparent methods for comparing CO$_2$ emissions across vehicle technologies. This paper proposes a machine learning--based framework for \textit{like-for-like} operational assessment of internal combustion engine vehicles (ICEVs) and electric vehicles (EVs) under identical, real-world driving conditions. The approach isolates technology-specific effects by holding the observed speed profile and environmental context fixed, enabling direct comparison of powertrain performance. Recurrent neural network models are trained independently for each domain to learn the mapping from contextual driving variables (speed, acceleration, temperature) to internal actuation variables (torque, throttle) and instantaneous CO\textsubscript{2}-equivalent emission rates. This structure allows the construction of counterfactual scenarios that answer: \textit{``What emissions would an EV have generated if it had followed the same driving profile as an ICEV?''} By aligning both vehicle types on a unified instantaneous emissions metric, the framework enables fair and reproducible evaluation of powertrain technologies. It offers a scalable foundation for credible, data-driven assessments of vehicle carbon performance under real-world operating conditions.
}


\jnlcitation{\cname{%
\author{Taylor M.},
\author{Lauritzen P},
\author{Erath C}, and
\author{Mittal R}}.
\ctitle{On simplifying ‘incremental remap’-based transport schemes.} \cjournal{\it J Comput Phys.} \cvol{2021;00(00):1--18}.}

\maketitle

\renewcommand\thefootnote{}
\footnotetext{\textbf{Abbreviations:} EV, electric vehicles, ICEV: internal combustion engine vehicles, MRV: measurement, reporting and verifications, EU-ETS: European Union Emissions Trading System, GHG: greenhouse gas RDC: real driving conditions, PEMS: portable emissions measurement system, ECU: electronic control unit, HVAC: heating, ventilation, air conditioning, RPM: rotation per minute, MAF: mass air flow, AFR: air fuel ratio, IQR: interquartile range, LSTM: Long short-term memory, RNN: recurrent neural network, MSE: mean square error.}

\renewcommand\thefootnote{\fnsymbol{footnote}}
\setcounter{footnote}{1}

\section{Introduction}\label{intro}
\input{introduction}

\section{System Model}
\input{model}

\section{Experimental Settings}
\input{experimental}

\section{Results and Discussion}

\input{results}

\section{Conclusion}

\input{conclusion}
\bibliography{wileyNJD-Vancouver}

\end{document}

%% file: introduction.tex
Decarbonizing road transport is essential to meeting climate targets. Road vehicles account for roughly one sixth of energy-related CO$_2$ emissions, so policy and investment choices must rely on assessments that are both credible and fair. Within this context, the transport sector is a critical arena for innovation and deployment, therefore accelerating the transition to a low‑carbon economy remains urgent~\cite{inproceedings}. Carbon markets are now widely used policy instruments and are increasingly intersecting with road transport~\cite{EC_ETS2_Overview,CPUC_LCFS,Wills_2025_Brazil_ETS}. The World Bank reports continuing expansion of carbon market pricing coverage, reinforcing the demand for comparable operational metrics in mobility, which raises the bar for measurement, reporting, and verification (MRV). In carbon markets, MRV denotes the end-to-end process that defines what is measured (including activity data and emission factors), specifies how results are reported in a consistent format, and requires independent checks to validate completeness and accuracy. Robust MRV underpins credit issuance and compliance by ensuring that comparable metrics are used across fuels, models, and technologies.

Globally, the share of emissions covered by direct carbon pricing keep rising, reinforcing the demand for consistent and comparable metrics in mobility~\cite{WorldBank2025StateTrendsCarbonPricing}. The new EU ETS2, the European Union's second emissions trading system, extends emissions trading to fuels for building and road transport, with phased MRV and compliance milestones, making upstream obligations depend on robust accounting of energy use that can be translated into CO\textsubscript{2}-equivalent under observed driving~\cite{EC_ETS2_Overview}. Established MRV guidance is converging as well: ISO 14083:2023 standardizes quantification and reporting of GHG emissions from transport chain operations, encouraging harmonized comparisons across modes and technologies~\cite{ISO14083_2023}. Jurisdictions already linking credits to carbon-intensity of delivered energy, e.g. California's Low Carbon Fuel Standard for electricity used in electric vehicles (EV) charging, also rely on defensible conversions from energy to CO\textsubscript{2}-equivalent to allocate benefits fairly~\cite{CPUC_LCFS}. In Brazil, the new regulated carbon market, the Brazilian Emissions Trading System, establishes a national framework whose sectoral coverage will be detailed in subsequent regulation~\cite{Wills_2025_Brazil_ETS}. Taken together, these developments increase the practical value of credible comparable CO\textsubscript{2}-equivalent estimates under observed driving, supporting, transparent accounting and better-informed decisions in transport policy and markets.

Progress now turns on methods that translate observed driving into comparable, accurate estimates of vehicles' energy use the resulting CO\textsubscript{2}-equivalent metrics. As electric cars have surpassed 17 million sales in 2024, over 20\% of global new-car registrations, stakeholders increasingly need operational metrics that hold up outside laboratories~\cite{IEA_GlobalEVOutlook_2025}. The persistent divergence between official type-approval figures and on-road performance reinforces the need for data-driven assessments grounded in actual driving conditions. Standardized tests are indispensable for regulation, but they cannot fully capture the dynamics of real driving conditions and by extension, provide the operational accounting required for decisions. For instance, the EU's first on-board fuel consumption monitoring report, covering almost 1 million vehicles first registered in 2021, confirms a persistent real-world vs. type-approval gap, reinforcing the value of operational assessments anchored in observed driving rather than laboratory cycles~\cite{EC_COM_2024_122_OBFCM}.

Reliable comparisons in real driving conditions (RDC) across powertrains hinge on evaluating them under the same observed driving context, because driver behaviour, route/traffic, temperature, and grade materially shift energy use and CO\textsubscript{2}-relevant outcomes. This is visible in controlled on-road campaigns and fleet-scale evidence. Early portable emissions measurement systems (PEMS) programs that tested multiple vehicles on the same reference routes (rural, urban, uphill/downhill, motorway) showed that average on-road results depend strongly on route design and conditions, and that vehicles driven on identical routes still exhibit sizable variability, hence the need to hold context constant when contrasting technologies~\cite{Weiss2011PEMS}. Complementary driver-effects experiments, e.g. 30 drivers piloting the same vehicle on the same route with PEMS, report significant between-driver dispersion in fuel use/emission, with worst-performing drivers consuming markedly more than others, regardless of experience~\cite{Huang2021Impact}. For EVs, long-horizon real-driving studies find that grade and ambient temperature strongly modulate specific energy consumption; for example, in one RDC-based study a +3\% uphill increased the specific energy consumption by about 50\% whereas a -3\% downhill decreased it by approximately 80\%, illustrating how uncontrolled terrain and conditions can swamp technology signals~\cite{Alwreikat2021Drivingbehaviour}. 

Head-to-head and route-matched studies operationalize a like-for-like comparison: vehicles are evaluated on identical itineraries (or on simulations constrained to the same route profile), so that the remaining differences more cleanly reflect the powertrain rather than route or driver. This is decisive for highlighting interpretable electric vehicles versus internal combustion engine vehicles (ICEV) contrasts. On a fixed 42-km urban itinerary in Erfurt (Germany), Braun and Rid (2017) instrumented a battery EV (BEV) and a ICEV and ran back-to-back drives with the same drivers, enabling paired comparisons under an identical route profile, an explicit like-for-like design~\cite{BraunRid2017_TRP_Erfurt}. A recent study conducted an EV-ICEV comparison under identical urban driving conditions at high altitude, constraining the EV analysis to the same route profile used for the gasoline vehicles~\cite{Puma2024ComparativeAnalysis}. Expanding beyond a single pair of vehicles, Suttakul et al. (2022) compared ICEV/hybrid-EV/plugin-hybrid-EV/battery-EV across defined route categories (city, rural, hill), showing how route characteristics systematically modulate relavite energy use and CO\textsubscript{2} outcomes, reinforcing the need for route-consistent comparisons when interpreting technologies effects~\cite{Suttakul2022_EnergyReports}. Complementary real-world validation work by Lee et al. (2024) contrasted chassis-dynamometer results with on-road measurements and reported 10.8-22.9\% disparities on the same route (and up to 29.3\% on different routes), arguing for practical testing under observed conditions~\cite{Lee2024_IJAT_RealWorldNecessity}. Taken together, these studies support evaluating different powertrains under the same observed driving context when the objective is a reliable technology comparison.

To capture such effects, modeling has evolved from established physics-informed approaches  \cite{das2024emissions} to modern data-driven techniques, with a strong focus on analyzing vehicle performance based on extensive real-world driving data \cite{lee2024energy} and applying deep learning to model instantaneous emissions \cite{yu2021novel}. Yet, a principled framework for a direct, counterfactual comparison remains an open methodological gap. Our work directly addresses this challenge.

This paper advances a ``like-for-like'' operational comparison procedure under a shared, observed driving scenario. The central idea is to (i) treat speed and other common contextual signals as the conditioning set; (ii) learn, within each domain, how those signals map to actuation and then to an instantaneous CO\textsubscript{2}-equivalent rate, in g/s; and (iii) assemble a counterfactual stream that answers: what CO\textsubscript{2}-equivalent rate would technology A register if it experienced exactly this technology-B velocity and context profile? Expressing both sides on the same instantaneous unit (g/s) avoids ad-hoc normalizations and enables pointwise as well as trip-level comparisons.

We illustrate the approach with a Brazilian case study. When electricity is the energy carrier, the conversion from instantaneous electrical power to a CO\textsubscript{2}-equivalent rate follows an official, grid-specific emission factor consistent with Brazil's national reporting; the corresponding fuel-based factors are applied when liquid fuels are the carrier.

Our contributions are threefold:
\begin{enumerate}
    \item A \textbf{fairness-preserving translation framework} that isolates technology effects under identical, real-world driving conditions;
    \item A \textbf{unified instantaneous emissions metric} that eliminates apples-to-oranges unit normalizations and aligns model targets across domains; and
    \item A \textbf{proxy-validation protocol} demonstrating that substituting model-predicted internal features for measured ones does not materially degrade emissions prediction, thereby supporting the final cross-domain comparison.
\end{enumerate}

We instantiate the approach on high-frequency datasets from both domains with a compact LSTM-based implementation; the framework, however, is model-agnostic.

\paragraph*{Paper structure.} Section~2 details the translation framework. Section~3 presents data and treatment. Section~4 describes the experimental setup. Section~5 reports results and ablations, and Section~6 discusses limitations and future work.

%% file: model.tex
Comparing emissions between internal combustion engine vehicles (ICEV) and electric vehicles (EV) is methodologically challenging. Even when two platforms traverse \emph{the same velocity profile}, their internal actuation, torque production, and the mapping from internal variables to emissions differs markedly. 
Although both powertrains expose internal actuation signals, their semantics are technology-specific. In ICE vehicles, `throttle' controls airflow (or an ECU torque proxy) and produces wheel torque only indirectly, through air-path dynamics, combustion control, and transmission state~\cite{KienckeNielsen2005}. The effective response depends on engine speed and gear because the engine must be steered across a \textit{brake specific fuel consumption} (BSFC) map with a relatively narrow high-efficiency region~\cite{Heywood2018}. In battery electric vehicles, the accelerator pedal is interpreted as a torque request that the inverter enforces directly at the motor, with near-instantaneous response across a broad high-efficiency region until the constant-power regime at higher speeds~\cite{Ehsani2010}. Similarly, the traction-torque channel differs by locus and dynamics: EV motor torque is measured/estimated at the machine and tightly controlled, whereas ICE `engine torque' is typically an ECU estimate and is filtered by gear ratios and shift logic before producing tractive effort~\cite{KienckeNielsen2005}. Because these channels differ in meaning, locus, and kinetics, they are treated as internal actuation signals used only within their respective domain models; cross-technology comparisons are performed on energy/emissions under identical observed driving context, not on the actuation signals themselves.

A fair comparison therefore requires a procedure that preserves the driving profile as a common reference while isolating technology-specific internal mechanisms. In our setting, differences in torque/actuator dynamics make naive feature-by-feature comparisons across domains invalid, even under identical speed trajectories (see Fig.~\ref{fig:training_pipeline}).

Our approach introduces a \emph{cross-domain translation} that factorizes the problem by domain and then recombines its components to yield counterfactual, yet technology-coherent, emission streams under a shared driving scenario. For each domain $D\in\{E,C\}$ (electric, combustion), we consider a pair of models: a \emph{feature model} $f_D$ that maps context variables common to both domains (e.g., speed, ambient/cabin temperatures, longitudinal acceleration) to the domain-specific internal actuation variables (hereafter, \emph{features}: traction torque and throttle); and an \emph{emissions model} $g_D$ that maps the triplet $[\text{speed},\text{torque},\text{throttle}]$ to emissions in that domain.
The selection of these context variables is grounded in established findings from
the literature. Sensitivity analyses have consistently demonstrated that vehicle
dynamics (speed, acceleration) and environmental factors are among the most 
significant predictors of an electric vehicle's energy demand and operational 
state \cite{asamer2016sensitivity}. In particular, environmental factors such as ambient 
temperature have a critical impact on powertrain efficiency and auxiliary loads (e.g., HVAC), making them indispensable for accurate energy estimation \cite{yi2017effects}.
By conditioning our model on these well-supported features, we ensure the framework's
physical plausibility.

It is this physically plausible representation that enables the cross-domain 
translation process. Translation arises by conditioning the feature model of one domain on the observed contextual trajectory and then feeding the resulting \emph{estimated} features into the same-domain emissions model. This poses an operationally grounded counterfactual: \emph{“what would the EV emit if it experienced exactly this ICEV velocity (and context) profile?”} The schematic with two symmetric branches (EV at left, ICEV at right) and the cross-domain flow is depicted in Fig.~\ref{fig:training_pipeline}.

We adopt the following notation to make the construction precise while keeping the exposition compact. Let a trip be indexed by $t=1,\dots,T$, with $\mathbf{x}_t\in\mathbb{R}^{p}$ the shared contextual variables (including speed $v_t$), $\mathbf{u}^{(E)}_t,\mathbf{u}^{(C)}_t\in\mathbb{R}^{2}$ the internal features (torque, throttle) for EV and ICEV, and $e^{(E)}_t,e^{(C)}_t\in\mathbb{R}$ the (instantaneous) emissions in g/s at time $t$. We model, within each domain, the pair of functions $(f_D,g_D)$ as:
\begin{subequations}\label{eq:domain_models}
\begin{align}
\text{(EV)}\quad & \mathbf{u}^{(E)}_t \approx f_{E}(\mathbf{x}_{1:t}), &
e^{(E)}_t &\approx g_{E}\!\big(v_t, \mathbf{u}^{(E)}_t\big),
\label{eq:ev_models}\\
\text{(ICEV)}\quad & \mathbf{u}^{(C)}_t \approx f_{C}(\mathbf{x}_{1:t}), &
e^{(C)}_t &\approx g_{C}\!\big(v_t, \mathbf{u}^{(C)}_t\big).
\label{eq:icev_models}
\end{align}
\end{subequations}

Given an \emph{observed} ICEV trajectory $(\mathbf{x}_{1:T},v_{1:T})$, the scheme infers the EV features that would obtain under the same conditions and then its corresponding emissions:
\begin{align}
\hat{\mathbf{u}}^{(E)}_{1:T} &= f_{E}(\mathbf{x}_{1:T}), \label{eq:ev_u_hat}\\
\hat{e}^{(E)}_{1:T} &= g_{E}\!\big(v_{1:T}, \hat{\mathbf{u}}^{(E)}_{1:T}\big). \label{eq:ev_e_hat}
\end{align}
For the same segment, the ICEV emissions (measured or estimated) are
\begin{equation}
\hat{e}^{(C)}_{1:T} = g_{C}\!\big(v_{1:T}, \mathbf{u}^{(C)}_{1:T}\big).
\label{eq:icev_e_hat}
\end{equation}
A pivotal ingredient is a \emph{proxy validation} that tests the reliability of within-domain substitution: replacing measured features with their model-based estimates should not materially degrade emissions prediction,
\begin{equation}
g_{D}\!\big(v_t, f_{D}(\mathbf{x}_{1:t})\big) \approx g_{D}\!\big(v_t, \mathbf{u}^{(D)}_t\big), 
\qquad D\in\{E,C\}.
\label{eq:proxy_validation}
\end{equation}

\medskip

\noindent\textbf{Emissions computation used in this study:} To facilitate a direct and equitable comparison between the two powertrain technologies, instantaneous emissions ($g/s$) were computed using methodologies appropriate to each \citep{Ryan2016EST,Faria2013RSER}.



\emph{\textbf{EV (electric-based)}.} For the electric vehicle (EV), a wheel-to-wheel approach was adopted to account for the carbon footprint of the electricity consumed \citep{Faria2013RSER}. The instantaneous power draw from the battery ($P_t$, in Watts), described as $P_t = |I_t V_t|$, is converted to an emission rate using a grid-specific emission factor ($\phi$) \citep{Ryan2016EST,SilerEvans2012EST}. The instantaneous emission rate for the EV, $e_t^{(E)}$, is thus calculated as:
\begin{equation}
e_t^{(E)} = \frac{P_t}{1000} \cdot \frac{\phi}{3600} \quad [g/s]
\label{eq:e_tE}
\end{equation}
The value employed in this study, $\phi = 38.5 \, \text{gCO\textsubscript{2}/kWh}$, represents the carbon intensity of electricity generation \citep{MCTI_SIRENE_2024_EN}. This factor is specifically representative of the emissions profile of Brazil's thermoelectric power generation, which is often dispatched to meet marginal demand \citep{SilerEvans2012EST}. Official data on the carbon intensity of the Brazilian national grid is periodically published by the Ministry of Science, Technology, and Innovation (MCTI) through its National Emissions Registry System (SIRENE) \citep{MCTI_SIRENE_2024_EN}.

\emph{\textbf{ICEV (fuel-based)}.} For the internal combustion engine vehicle (ICEV), emissions are calculated directly from fuel consumption \citep{Frey2003JAWMA}. The methodology considers the vehicle's instantaneous efficiency ($K_t$ in km/L), its velocity ($v_t$), the volumetric percentage of ethanol in the fuel blend ($P$), and the distinct emission factors for gasoline ($F_g$) and ethanol ($F_e$) \citep{SuarezBertoa2015Fuel,Karavalakis2012Fuel}. The instantaneous emission rate for the ICEV, $\dot{e}_t^{(C)}$, is given by:
\begin{equation}
\dot{e}_t^{(C)} = \frac{v_t}{3600 K_t} \left[ \left(1 - \frac{P}{100}\right) F_g + \frac{P}{100} F_e \right] \quad [g/s]
\label{eq:icev_edot_t}
\end{equation}

The study uses emission factors of $F_g = 2310 \, g/L$ for gasoline and $F_e = 1510 \, g/L$ for ethanol. These values are derived from the stoichiometric combustion of the fuels and are consistent with findings and practices \textbf{reported in emissions studies with ethanol-gasoline blends}\citep{Karavalakis2012Fuel,SuarezBertoa2015Fuel}.

The framework’s fairness criterion then compares $\dot{e}^{(E)}_t$ and $\dot{e}^{(C)}_t$ under the same velocity trajectory and shared context, while technology-specific internal mechanisms remain encapsulated within $(f_D,g_D)$. The factorization is model-agnostic; any sequence model capable of capturing nonlinear temporal dependencies can instantiate $f_D$ and $g_D$. For concreteness in this study we adopt LSTM-based networks with a fixed architecture: one LSTM layer (32 hidden units, batch-first), followed by a fully connected layer with 32 ReLU units and a linear output head (one output for emissions; two outputs for torque and throttle). Inputs are MinMax-scaled to $[0,1]$, time series are segmented into sliding windows of length $10$, and optimization uses MSE with Adam and a cosine learning-rate scheduler with warm-up.

%% file: experimental.tex
This section describes the experimental procedure adopted to compare CO2 emissions between an internal combustion engine vehicle (ICEV) and an electrical vehicle (EV). 
Although both vehicles can perform the same velocity trajectory, their torque dynamics and accelerator actuation differ substantially, which makes a direct comparison of features impossible. In order to overcome this asymmetry, a three-stage flow was carried out -- Training, Validation and Test-Time --, in which models are separately trained in each domain and integrated only when necessary, preserving comparability under the same speed profile.

\subsection{Data}
We use two complementary sources, one for each powertrain domain. 
\textbf{EV domain:} a high-frequency publicly available dataset with measurements of vehicle speed ($v_t$, km/h), traction motor torque (Nm), accelerator throttle (\%), and contextual variables shared with the combustion domain (ambient/cabin temperatures, longitudinal acceleration). Instantaneous EV emissions are not measured; instead they are \emph{derived} from electrical power as detailed in Eq.~\eqref{eq:e_tE}~\cite{IEEEevDataset}. 
\textbf{ICEV domain:} a chassis-dyno dataset covering multiple trips for different vehicles, with analogous variables (speed, torque/tractive force, throttle or fuel proxies, and thermal context), plus fuel-related signals enabling computation of instantaneous ICEV emissions via Eq.~\eqref{eq:icev_edot_t} (ethanol share $P$ in \%, efficiency $K_t$ in km/L, and emission factors $F_g=2310~\mathrm{g/L}$ and $F_e=1510~\mathrm{g/L}$), the raw data is accessible at~\cite{ANL_D3_Conventional_2025}.

Across both domains, we standardize symbol conventions and units for downstream modeling: speed in km/h, torque in Nm, longitudinal acceleration in m/s$^2$, temperatures in $^\circ$C, and emissions as instantaneous rates in g/s. The modeling tasks use the shared-context variables as inputs for feature translation and the triplet $[v_t,\text{torque},\text{throttle}]$ for the emission models, ensuring that cross-domain comparisons are conditioned on the same driving scenario.

\subsection{Data Treatment}
We apply a uniform preprocessing pipeline to harmonize time bases, units, and nomenclature, followed by vehicle-specific steps. Unless stated otherwise, all filtering is performed prior to feature/target construction.
\begin{enumerate}
    \item \textbf{Integrity checks and filtering.} Removal of records with invalid or non-physical values (e.g., negative timestamps, negative tractive force/engine torque where not expected), inconsistent identifiers, or corrupted trip metadata. 
    \item \textbf{Time-base handling.} When needed, resampling to a uniform time grid and forward-filling short gaps in identifiers to preserve trip continuity, while avoiding any smoothing that would bias instantaneous emissions.
    \item \textbf{Unit harmonization and renaming.} Conversion to SI/metric units (e.g., mph$\to$km/h; volume-flow to mass-flow where applicable) and adoption of a consistent variable schema across EV/ICEV (speed, torque, throttle, temperatures, longitudinal acceleration, and emissions).
    \item \textbf{Derived features.} Construction of throttle proxies when the pedal signal is not directly available; computation of motor torque from tractive force and wheel radius where needed; estimation of longitudinal acceleration from speed differences; and standardization of contextual temperatures.
    \item \textbf{Emission targets.} Instantaneous emissions (in g/s) are computed strictly following the equations:
    EV by the electric-based rate in Eq.~\eqref{eq:ev_e_hat}, and ICEV by the fuel-based rate in Eq.~\eqref{eq:icev_edot_t}. 
    This alignment ensures that $g_D$ learns and predicts targets in the same units across domains.
\end{enumerate}

\paragraph*{Vehicle-specific notes.}
\textbf{Infiniti QX50 (ICEV, CVT).}
(i) \emph{File ingestion \& IDs.} We start from a single combined file produced from the source \texttt{.txt} files whose headers are validated against the expected schema; incompatible files are discarded. \texttt{Trip} identifiers are taken from the source filenames (with \texttt{Test\_ID} forward filled where present to preserve continuity), and finally normalized to a numeric format (removing suffixes such as ``Test Data'').\\
(ii) \emph{Integrity checks.} Removal of records with \texttt{Time}<0; within each \texttt{Trip}, exclusion of samples with $\Delta t\le 0$ and duplicate timestamps. When available, negative engine torque (\texttt{Eng\_torque\_TCM}<0) is also excluded. No smoothing is applied.\\
(iii) \emph{Unit harmonization \& renaming.} Speed is converted from mph$\rightarrow$km/h and variables are mapped to the unified schema: \texttt{Throttle [\%]} (from \texttt{Pedal\_accel\_CAN2\_per} when needed), \texttt{Motor Torque [Nm]}, \texttt{Ambient/Cabin/Heat Exchanger Temperature [°C]}, \texttt{Time [s]}, \texttt{Velocity [km/h]}. Drivetrain rotation channels (RPM) are preserved and renamed for readability:
\begin{itemize}
  \item \texttt{Eng\_spd\_CAN2\_\_rpm} $\rightarrow$ 
  \item \texttt{Input\_CVT\_Shaft\_Rev\_TCM\_\_rpm} 
  \item \texttt{Trans\_primary\_pulley\_speed\_CAN2\_\_rpm} 
  \item \texttt{Trans\_secondary\_pulley\_speed\_CAN2\_\_rpm} 
  \item \texttt{Trans\_slip\_speed\_TCM\_\_rpm} $\rightarrow$ 
\end{itemize}
(iv) \emph{Derived features.} Longitudinal acceleration $a_t$ is computed by finite differences per \texttt{Trip} using the true (possibly irregular) time base, $a_t=\Delta v / \Delta t$, with each trip's first sample set to $0$\,m/s$^2$ and without any smoothing. We also compute \texttt{Wheel RPM [rpm]} from vehicle speed and the assumed 19'' wheel/tire radius:
\begin{equation}
\mathrm{WheelRPM} \approx 7.150\, v_{\mathrm{km/h}},
\label{eq:wheelrpm}
\end{equation}

If the exact tire size is available, $R$ is updated accordingly.\\
(v) \emph{Emission target (ICEV).} Instantaneous CO$_2$ [g/s] is derived from fuel flow: preferentially from MAF using the stoichiometric AFR (14.7) and fuel density (740\,g/L), otherwise from \texttt{Eng\_FuelFlow\_Direct2 [gps]} ($\rightarrow$ L/s) or \texttt{Eng\_FuelFlow\_Direct [ccps]} ($\rightarrow$ L/s). The km/L ratio is $K=v_{\text{km/h}}/\dot{V}_{\text{fuel}}$; composition constants are then applied and the result mapped to the instantaneous rate per Eq.~\eqref{eq:icev_edot_t}, keeping units aligned with other vehicles.\\
(vi) \emph{Strict filtering (variant).} For the \texttt{QX50 strict} set, in addition to (ii)--(v) we apply: (a) global physical bounds on key variables (speed 0--250\,km/h; \texttt{Throttle} 0--100\%; \texttt{Motor Torque} 0--1200\,Nm; plausible temperature ranges; CO$_2>0$); (b) removal of records with $|a_t|>10$\,m/s$^2$; (c) global IQR outlier removal with $k=3$ (no winsorization).\\

\textbf{Chevrolet Blazer (ICEV).} 
(i) Forward-fill of missing \texttt{Test\_ID} values; removal of rows with unresolved \texttt{Test\_ID}. 
(ii) Exclusion of records with negative time (\texttt{Time}<0) or negative engine torque (\texttt{Eng\_torque\_TCM}<0). 
(iii) Removal of corrupted trips (e.g., \texttt{Trip}=61177923.13997565). 
(iv) Fuel flow (L/h) computed from \texttt{Eng\_MAF\_total\_ECM} using the stoichiometric air–fuel ratio (14.7) or, when available, taken from \texttt{Eng\_FuelFlow\_Direct}. 
(v) CO$_2$ emissions obtained from fuel flow and fuel composition constants, then mapped to instantaneous rate via Eq.~\eqref{eq:icev_edot_t}. 
(vi) Speed converted from mph to km/h, and variables renamed to the unified schema; rows with missing values in the analysis set removed.

\textbf{Chrysler Pacifica (ICEV).}
(i) Input validation: presence of essential variables (\texttt{Time}, \texttt{Dyno\_Spd}, \texttt{Cell\_Temp}, \texttt{Radiator\_Air\_Outlet\_Temp}, \texttt{Cabin\_Temp}, \texttt{Dyno\_TractiveForce}, and at least one fuel measurement). 
(ii) Definition of \texttt{Trip} identifiers from source file names. 
(iii) Removal of negative time (\texttt{Time}<0) and negative tractive force (\texttt{Dyno\_TractiveForce}<0); when available, exclusion of negative engine torque (\texttt{Eng\_torque\_TCM}<0). 
(iv) Throttle position [\%] estimated as a normalized function of \texttt{Eng\_FuelFlow\_Direct}. 
(v) Motor torque [Nm] computed as $T=\texttt{Dyno\_TractiveForce}\times R$ with wheel radius $R=0.3$\,m. 
(vi) Longitudinal acceleration [m/s$^2$] obtained from speed differences over time. 
(vii) CO$_2$ volume flow converted from m$^3$/min to m$^3$/s and then to mass emissions [g/s] using gas density and dilution concentration; final instantaneous rate follows Eq.~\eqref{eq:icev_edot_t}. 
(viii) Speed converted from mph to km/h. 
(ix) Outlier filtering: records with motor torque $\geq 400$\,Nm or CO$_2>25$\,g/s removed. 
(x) Variables renamed to the unified schema; rows with missing values in the selected columns removed.

\textbf{BMW i3 (EV).}
(i) Selection of relevant measurement channels from the original $\sim$50 recorded signals, retaining \texttt{Time~[s]}, \texttt{Velocity~[km/h]}, \texttt{Throttle~[\%]}, \texttt{Motor Torque~[Nm]}, \texttt{Ambient Temperature~[°C]}, \texttt{Cabin Temperature Sensor~[°C]}, \texttt{Heat Exchanger Temperature~[°C]}, and \texttt{Longitudinal Acceleration~[m/s$^2$]}.  
(ii) Creation of a \texttt{Trip} identifier from source file names for the 72 real-world driving sessions.  
(iii) Removal of records with missing values in the selected variables.  
(iv) Computation of instantaneous CO$_2$ emissions [g/s] from electric power $P_t = |V_{\text{batt}} I_{\text{batt}}|$ using the energy-to-emission conversion factor $\phi=38.5$\,g/Wh, strictly following Eq.~\eqref{eq:e_tE}; regenerative braking phases (negative current) were excluded from the computation.  
(v) Standardization of variable names and units to match the unified schema and export to harmonized CSV format.
ation of variable names and units to match the unified schema and export to harmonized CSV format.

\medskip
All the steps above yield harmonized, high-frequency series with consistent units and symbols for both domains, enabling the translation-based framework to learn $(f_D,g_D)$ with emission targets defined as instantaneous rates  and to compare technologies under identical driving conditions.

\subsection{Model Architecture and Training Protocol}

We implement all models in PyTorch using a recurrent neural network (RNN) architecture based on Long Short-Term Memory (LSTM) units, chosen for their ability to capture temporal dependencies in time-series data. The network architecture is somewhat fixed across all tasks and consists of: (1) a one or two-layer LSTM with 32 (EV domain) or 64 (ICEV) hidden units (configured as batch-first); (2) a fully connected layer with 32/64 neurons, using ReLU activation; and (3) a final linear output layer. The number of output neurons is adapted to the task: one for emissions models ($g_D$) and two for feature models ($f_D$, predicting torque and throttle).

Input time series are preprocessed by segmenting them into sliding windows of 10 timesteps to model short-term dynamics, and normalizing features to the interval $[0, 1]$ using Min-Max scaling. Models are trained to minimize the Mean Squared Error (MSE) loss function using the Adam optimizer, coupled with a cosine learning rate scheduler with warm-up to ensure stable convergence. The number of training epochs is adjusted based on the complexity of each task, where for the EV domain the number of was set 20 epochs, for the ICEV domain we adopted 50 epochs. All training was conducted on an NVIDIA A100-SXM4-40GB GPU.

\subsection{Three-Stage Comparative Framework}

\begin{figure*}[!htbp]
    \centering
    \includegraphics[width=0.97\textwidth]{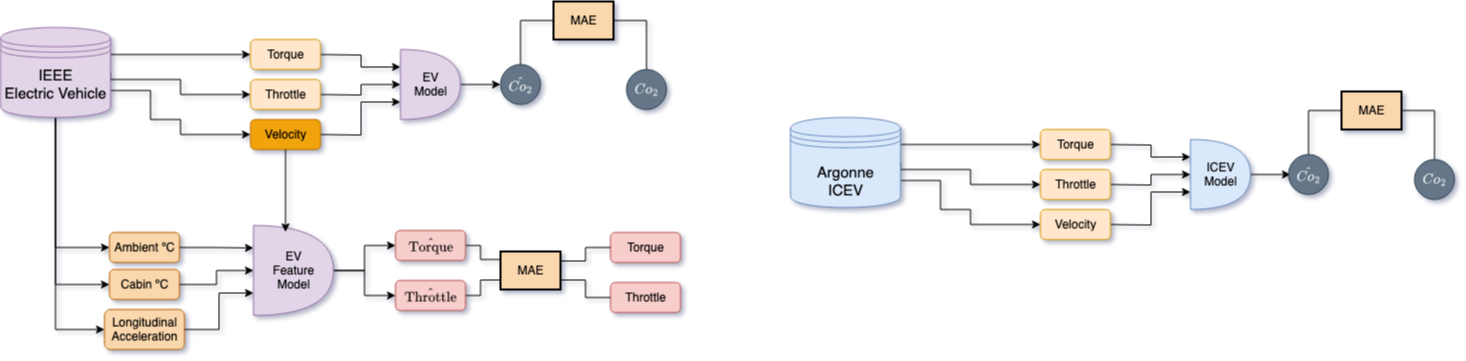}
    \caption{Stage 1: Domain-specific training pipeline. For each domain (EV and ICEV), an Emissions Model and a Feature Model are trained independently.}
    \label{fig:training_pipeline}
\end{figure*}

\textbf{Stage 1: Domain-Specific Training}
As depicted in Figure~\ref{fig:training_pipeline}, the training process is performed independently within each domain ($D \in \{\text{EV, ICEV}\}$), which two specialized models could be trained:
\begin{itemize}
    \item \textbf{Emissions Model ($g_D$):} This model learns to map the vehicle's internal state variables, specifically $[\text{velocity, torque, throttle}]$, to its instantaneous $CO_2$ emissions ($g/s$).
    \item \textbf{Feature Model ($f_D$):} This model learns to predict the domain-specific actuation variables, $[\text{torque, throttle}]$, using only contextual variables common to both domains, namely: velocity, ambient temperature, cabin temperature, and longitudinal acceleration.
\end{itemize}
This decomposition is the cornerstone of our framework, as it separates the universal driving context from the technology-specific vehicle response. Despite that, in this study we only train $f_D$ for EV's, as detailed as follows. 

\textbf{Stage 2: Proxy Validation}

\begin{figure*}[!htbp]
    \centering
    \includegraphics[width=0.6\textwidth]{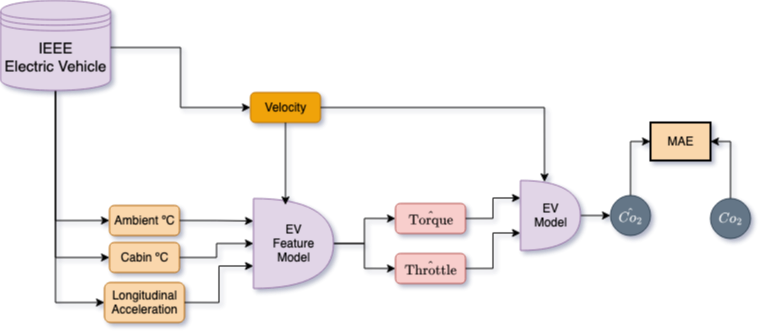}
    \caption{Stage 2: Proxy validation pipeline. The performance of the emissions model is tested when its inputs (torque, throttle) are supplied by the feature model, quantifying the error introduced by the proxy.}
    \label{fig:validation_pipeline} 
\end{figure*}

Before any cross domain comparison, we validate robustness within each domain (Figure~\ref{fig:validation_pipeline}). 
First, we assess the baseline performance of the emissions models $g_{\text{EV}}$ and $g_{\text{ICEV}}$ using measured ground truth. Secondly, we carry out a proxy validation to assess the impact of using predicted features ($f_{EV}$) -- as opposed as measured torque and throttle --,  as input to the emission model ($g_{EV}$). This enabled to quantify the prediction error in relation to the $CO_2$ emissions ground-truth, i.e., when using predicted features in contrast to measured ones. 

As the EV embedded counterfactual $\hat e^{(E)}=g_{\text{EV}}\!\big(v^{(C)},\,f_{\text{EV}}(x^{(C)})\big)$ has no paired EV ground truth for the identical ICEV trajectory (no simultaneous or replicated EV run), a direct check is not available, therefore, a within-domain evaluation was necessary. Besides that, the Proxy validation serves as evidence that the required substitution does not materially degrade accuracy; a direct counterfactual check would require paired EV–ICEV data collected on the same route and conditions.





\textbf{Stage 3: Proposed Test-Time Counterfactual Analysis}
In this final stage, the EV can be treated as a counterfactual system under identical operating conditions as an ICEV's trajectory. The ICEV context (velocity, temperatures, longitudinal acceleration) would be fed to the pre trained EV Feature model ($f_E$) to infer the torque and throttle that an EV would likely produce. These inferred signals, together with the velocity profile, would then be passed to the EV Emissions model ($g_E$) to generate the counterfactual EV emissions series. 

Although this stage is not executed in the present study, we believe to establish its viability by showing that the proxy substitution inside $g_E$ produces negligible degradation and that the learned components reach low error levels. In other words, our findings suggest that rather than a methodological limitation, the inverse map from the shared context to engine actuation is weakly identifiable for a CVT without transmission-state/lock-up information. Besides that, the results we present—bounded error for $f_E$ on shared inputs and negligible loss when $\hat u^{(E)}$ replaces measured actuation inside $g_E$ (Table~\ref{tab:ev_mae}, Figures~\ref{fig:ev_feature_test}–\ref{fig:ev_emission_validation})—demonstrate that, under the same preconditions on the ICEV side (e.g., explicit transmission state or a non-CVT powertrain), the same cross domain translation is feasible in the opposite direction. Moreover, we believe these results provide the reliability conditions required to run the counterfactual analysis over ICEV trips in future applications.

%% file: results.tex
This section deepens the analysis of the three-stage procedure (training, proxy validation, and test-time counterfactual) and discusses what the numbers imply for cross-technology comparison. We first examine ICEV results, then the EV branch (including proxy validation), and finally comment on readiness for counterfactual application. All emissions are instantaneous rates (g/s).

\subsection{ICEV}
\label{sec:results_icev}

\begin{figure}[!t]
    \centering
    \includegraphics[width=0.95\linewidth]{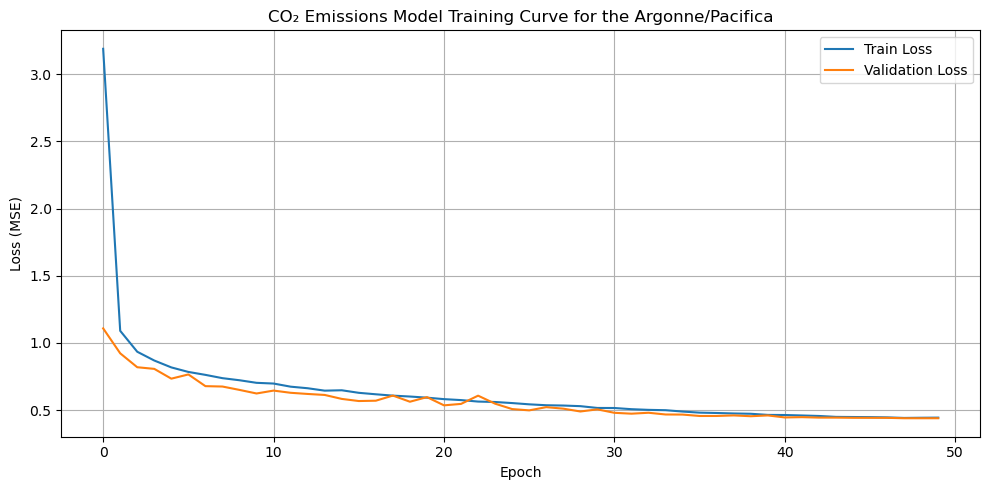}
    \caption{ICEV emissions model training/validation loss for \textit{Chrysler Pacifica}. Loss (MSE) vs.\ epoch shows rapid convergence and stable generalization. Final MAE $\approx$ 0.300, MSE $\approx$ 0.497 (validation set).}
    \label{fig:icev_pacifica_training}
\end{figure}

\begin{figure}[!t]
    \centering
    \includegraphics[width=0.95\linewidth]{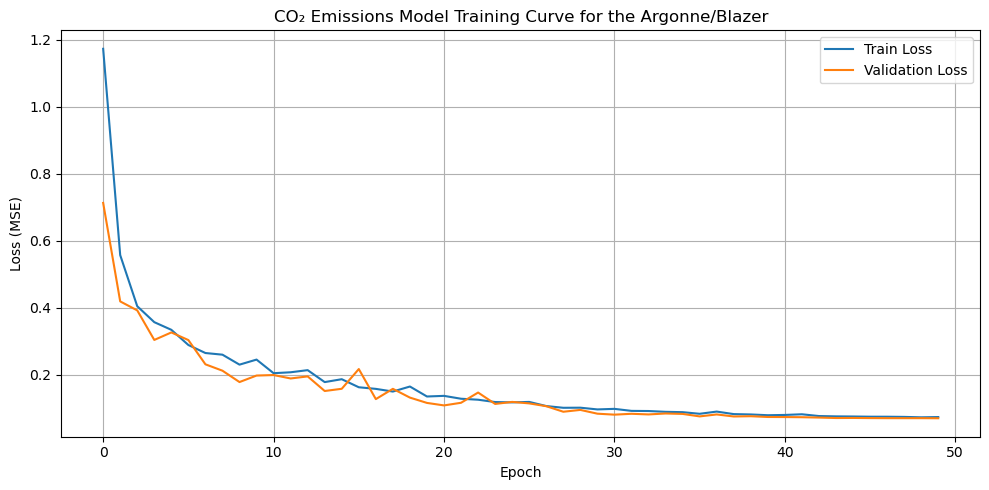}
    \caption{ICEV emissions model training/validation loss for \textit{Chevrolet Blazer}. Training and validation curves remain tightly coupled after the initial transient. Final MAE $\approx$ 0.153, MSE $\approx$ 0.072 (validation set).}
    \label{fig:icev_blazer_training}
\end{figure}

\begin{figure}[!t]
    \centering
    \includegraphics[width=0.95\linewidth]{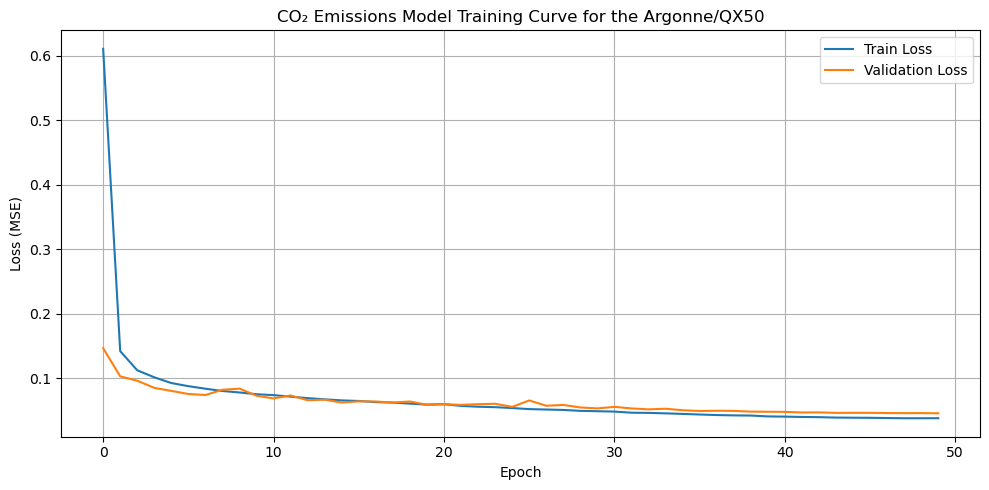}
    \caption{ICEV emissions model training/validation loss for \textit{Infiniti QX50}. The model stabilizes after the early epochs without signs of overfitting. Final MAE $\approx$ 0.097, MSE $\approx$ 0.046 (validation set).}
    \label{fig:icev_qx50_training}
\end{figure}

Across the three ICEV platforms, training and validation losses fall rapidly and then stabilize without pronounced overfitting, which indicates that a short context window (10 steps) suffices to capture the local dynamics that drive instantaneous CO$_2$ on the chassis-dyno data (see Figures~\ref{fig:icev_pacifica_training}--\ref{fig:icev_qx50_training}). In practice, most of the predictability for pointwise emissions resides in short-horizon signals (speed/actuation), and a single-layer LSTM is adequate for this task.

\begin{figure*}[!t]
    \centering
    \includegraphics[width=0.49\textwidth]{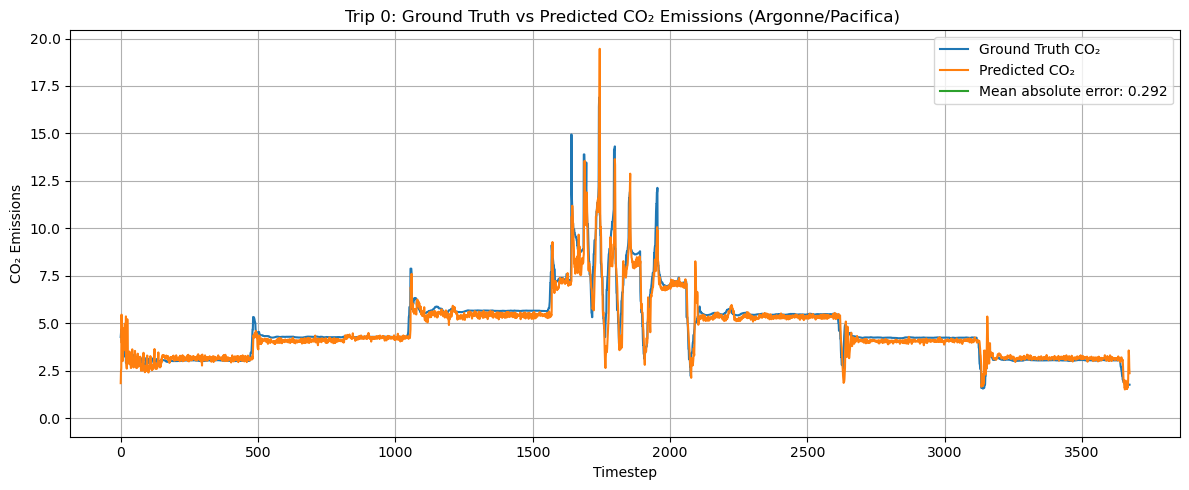}
    \includegraphics[width=0.49\textwidth]{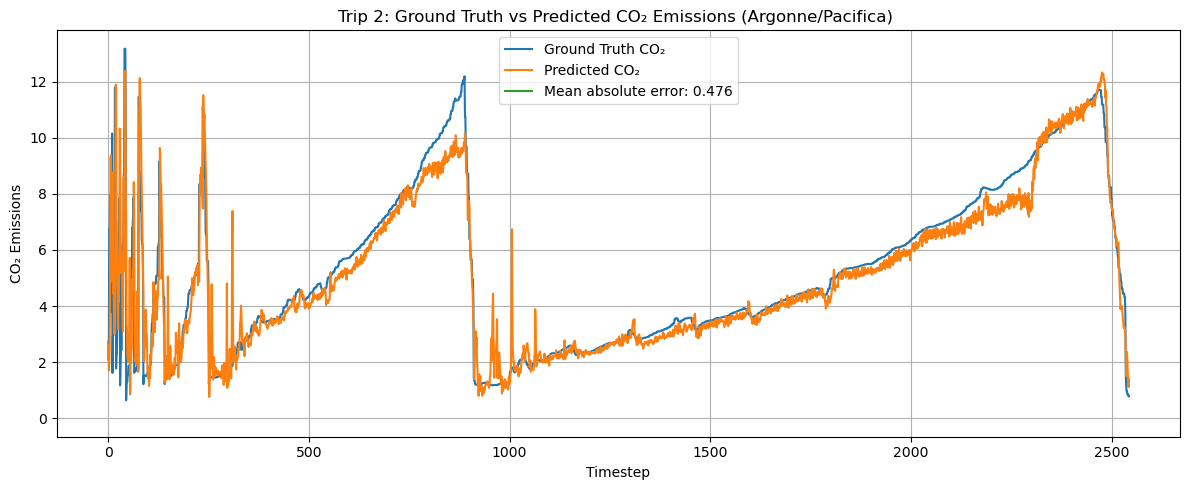}
    \caption{ICEV test-set best (left) vs.\ worst (right) cases for the \textit{Chrysler Pacifica} emissions model.}
    \label{fig:icev_pacifica_cases}
\end{figure*}

\begin{figure*}[!t]
    \centering
    \includegraphics[width=0.49\textwidth]{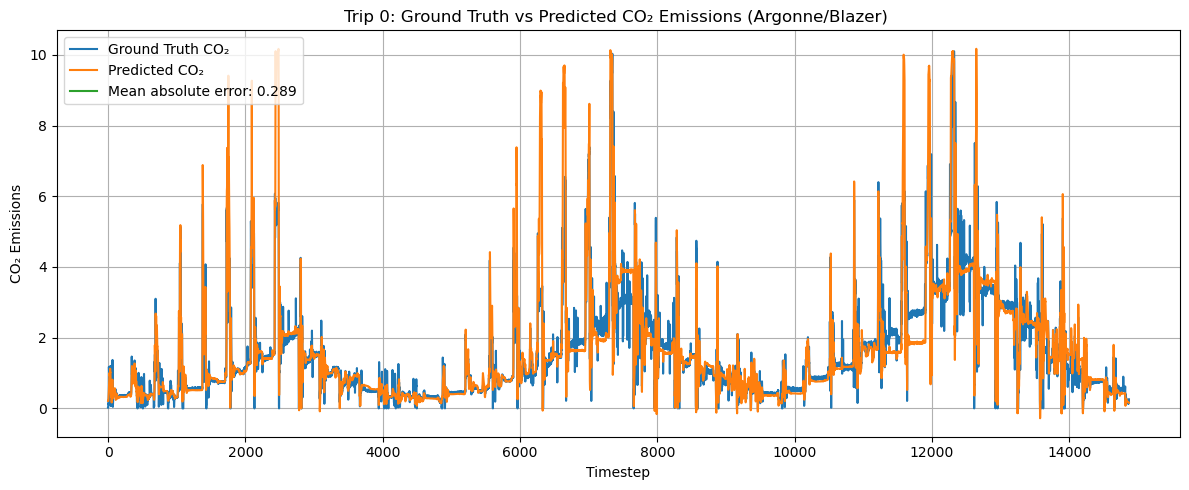}
    \includegraphics[width=0.49\textwidth]{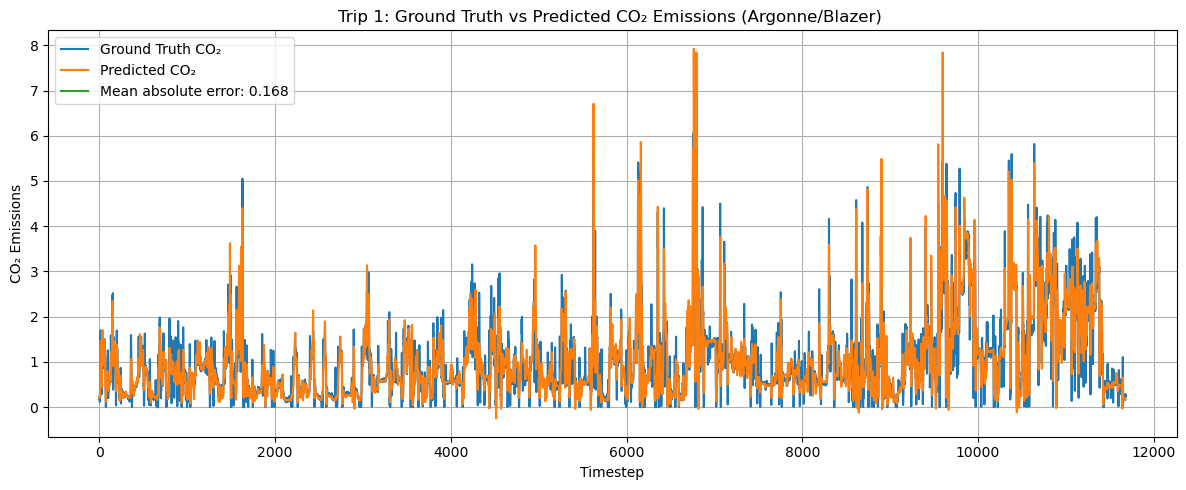}
    \caption{ICEV test-set best (left) vs.\ worst (right) cases for the \textit{Chevrolet Blazer} emissions model.}
    \label{fig:icev_blazer_cases}
\end{figure*}

\begin{figure*}[!t]
    \centering
    \includegraphics[width=0.49\textwidth]{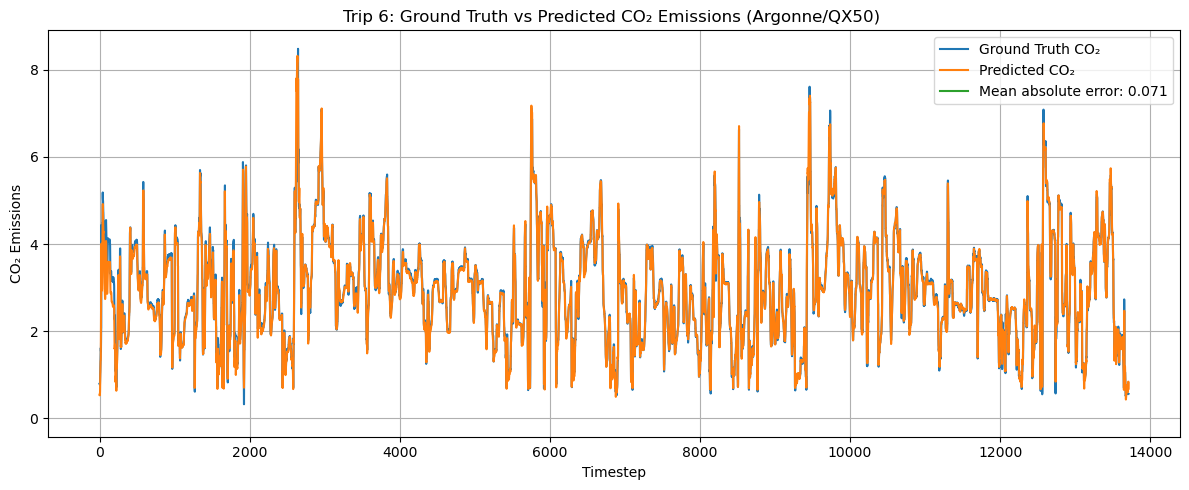}
    \includegraphics[width=0.49\textwidth]{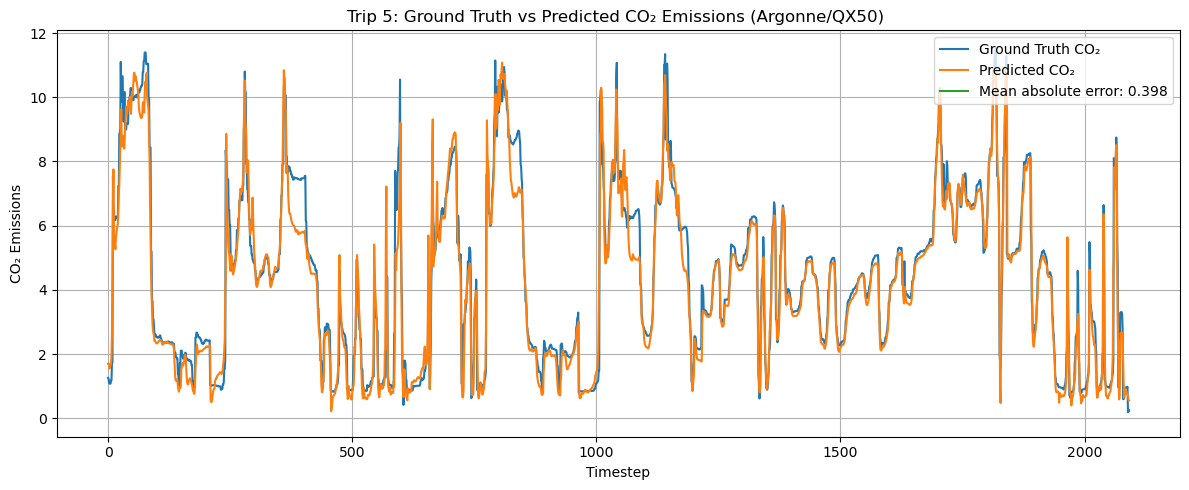}
    \caption{ICEV test-set best (left) vs.\ worst (right) cases for the \textit{Infiniti QX50} emissions model.}
    \label{fig:icev_qx50_cases}
\end{figure*}

Tables~\ref{tab:icev_qx50_mae}--\ref{tab:icev_pacifica_mae} summarize held-out MAE by trip. For the \textit{QX50} (7 trips) the mean/median/min/max are 0.254/0.252/0.071/0.398 g/s. For the \textit{Blazer} (2 trips), mean/median are 0.2285/0.2285 g/s. For the \textit{Pacifica} (4 trips) the mean/median/min/max are 0.359/0.335/0.292/0.476 g/s. Aggregating all 13 ICEV trips yields mean 0.283 g/s, median 0.289 g/s, min 0.071 g/s, max 0.476 g/s. Beyond centrality, dispersion is moderate: the aggregated interquartile range is roughly $[0.247,\,0.317]$ g/s and the sample standard deviation is $\approx 0.101$ g/s (computed from Tables~\ref{tab:icev_qx50_mae}--\ref{tab:icev_pacifica_mae}), indicating non-negligible trip-to-trip variability.

\begin{table}[!t]
  \centering
  \caption{\textit{Infiniti QX50} — Mean Absolute Error (MAE) on test trips.}
  \label{tab:icev_qx50_mae}
  \small
  \setlength{\tabcolsep}{3.5pt}
  \begin{tabularx}{\columnwidth}{@{}r>{\centering\arraybackslash}X@{}}
    \toprule
    Trip & CO$_2$ Emissions (MAE, g/s) \\
    \midrule
    0 & 0.247 \\
    1 & 0.235 \\
    2 & 0.317 \\
    3 & 0.252 \\
    4 & 0.259 \\
    5 & 0.398 \\
    6 & 0.071 \\
    \bottomrule
  \end{tabularx}
\end{table}

\begin{table}[!t]
  \centering
  \caption{\textit{Chevrolet Blazer} — Mean Absolute Error (MAE) on test trips.}
  \label{tab:icev_blazer_mae}
  \small
  \setlength{\tabcolsep}{3.5pt}
  \begin{tabularx}{\columnwidth}{@{}r>{\centering\arraybackslash}X@{}}
    \toprule
    Trip & CO$_2$ Emissions (MAE, g/s) \\
    \midrule
    0 & 0.289 \\
    1 & 0.168 \\
    \bottomrule
  \end{tabularx}
\end{table}

\begin{table}[!t]
  \centering
  \caption{\textit{Chrysler Pacifica} — Mean Absolute Error (MAE) on test trips.}
  \label{tab:icev_pacifica_mae}
  \small
  \setlength{\tabcolsep}{3.5pt}
  \begin{tabularx}{\columnwidth}{@{}r>{\centering\arraybackslash}X@{}}
    \toprule
    Trip & CO$_2$ Emissions (MAE, g/s) \\
    \midrule
    0 & 0.292 \\
    1 & 0.365 \\
    2 & 0.476 \\
    3 & 0.304 \\
    \bottomrule
  \end{tabularx}
\end{table}

Event-level behavior on the test set is illustrated by the best– and worst–case trips for each ICEV (Figures~\ref{fig:icev_pacifica_cases}--\ref{fig:icev_qx50_cases}). For the \textit{Chrysler Pacifica} (Figure~\ref{fig:icev_pacifica_cases}), the best case shows close tracking of the ground truth across load changes, with small residuals outside brief spikes; in the worst case, errors concentrate in long monotonic ramps and isolated peaks, where the model tends to underestimate the peak magnitude and to lag during the fastest accelerations. This is consistent with the use of a throttle \emph{proxy} for Pacifica and with the general difficulty of capturing very high slew events with short sequence windows. For the \textit{Chevrolet Blazer} (Figure~\ref{fig:icev_blazer_cases}), the best case exhibits smoother, lower-‑amplitude demand and the model follows the envelope well, whereas the worst case presents dense bursts of short spikes that are systematically underpredicted and occasionally phase-‑shifted; the contrast aligns with Table~\ref{tab:icev_blazer_mae} (0.168 vs.\ 0.289~g/s). For the \textit{Infiniti QX50} (Figure~\ref{fig:icev_qx50_cases}), both best and worst trips include frequent high‑-frequency peaks superimposed on moderate baselines; residuals grow around the sharpest bursts but the base-‑level alignment remains strong, which is coherent with the trip-‑level dispersion seen in Table~\ref{tab:icev_qx50_mae} and with the variability expected from a CVT powertrain.

Two plausible contributors to the differences across ICEV platforms are signal availability/proxies and powertrain particulars. For the Pacifica, throttle is derived from fuel‑flow proxies instead of a direct pedal signal, which may inject additional noise and help explain its higher MAE distribution (see Data Treatment for Pacifica). In turn, the QX50 uses a CVT; while the emissions model does not attempt to infer transmission state, variability in torque delivery across trips can still affect the mapping to instantaneous CO$_2$. These factors likely drive the observed dispersion and should motivate targeted ablations (e.g., re‑training with enriched actuation signals when available).

\subsection{EV (proxy validation and counterfactual readiness)}
\label{sec:results_ev}

\begin{figure*}[!t]
    \centering
    \includegraphics[width=0.49\textwidth]{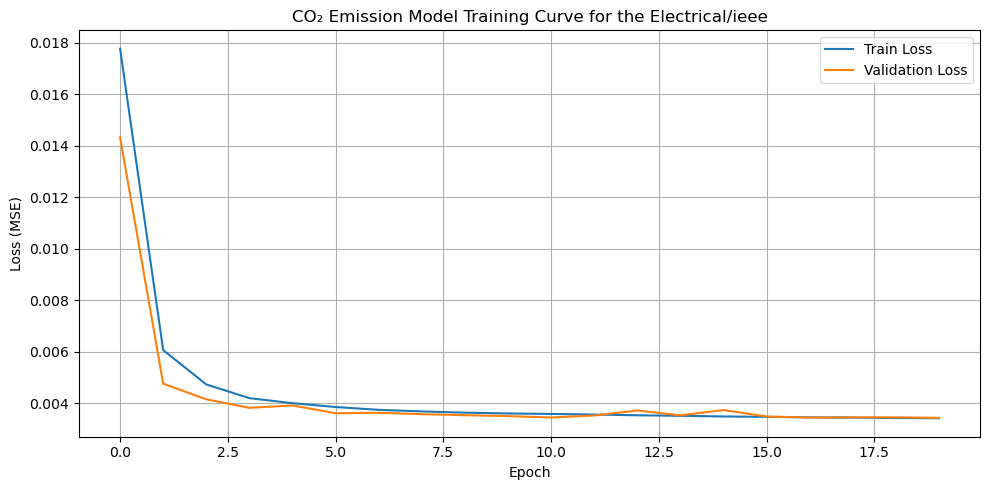}
    \includegraphics[width=0.49\textwidth]{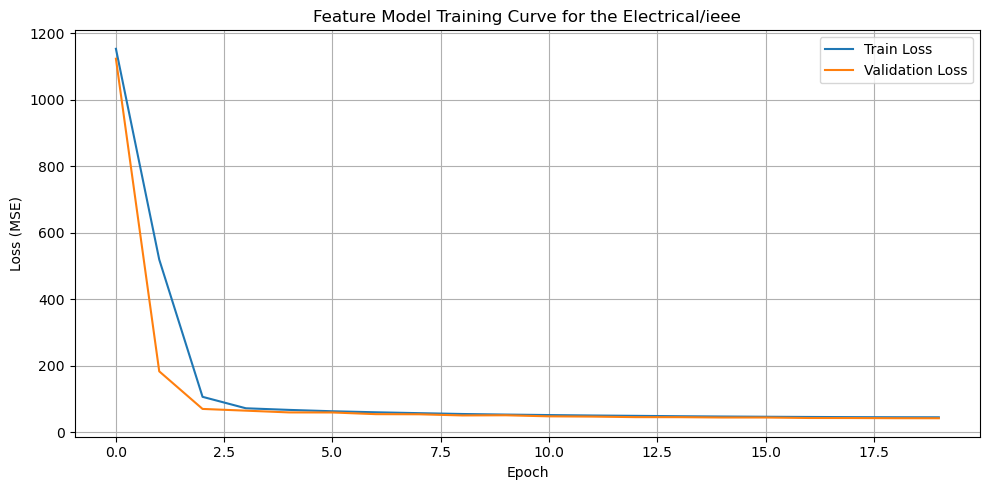}
    \caption{Training curves (EV). Left: EV-Emissions (final MAE $\approx$ 0.028, MSE $\approx$ 0.00344). Right: EV-Feature (final MAE $\approx$ 3.74, MSE $\approx$ 41.96).}
    \label{fig:ev_training_curves}
\end{figure*}

In the EV domain (BMW i3), the EV-Emissions model converges to low error (final MAE $\approx 0.028$, MSE $\approx 0.00344$) and the EV-Feature model reaches MAE $\approx 3.74$, MSE $\approx 41.96$ in natural units (Nm, pp); see Figure~\ref{fig:ev_training_curves}. This indicates that the EV branch has learned stable mappings for both emissions and actuation, enabling composition at test time.

Table~\ref{tab:ev_mae} shows that substituting measured $(\text{torque},\text{throttle})$ with EV-Feature predictions induces negligible degradation inside EV-Emissions: the median MAE is 0.0288~g/s (direct) vs.\ 0.0290~g/s (proxy); the per-trip difference (proxy $-$ direct) has median $\approx -0.0025$~g/s, interquartile range $\approx [-0.0066,\;0.0002]$~g/s, and range $[-0.013,\;0.004]$~g/s. In 10/14 trips the proxy MAE is $\le$ the direct MAE, suggesting a mild denoising effect from the feature model and reinforcing counterfactual viability.

\begin{table}[!t]
  \centering
  \caption{\textit{BMW i3} — Mean Absolute Error (MAE) across held-out trips. ``CO$_2$'' uses measured actuation; ``Proxy CO$_2$'' feeds EV-Feature predictions.}
  \label{tab:ev_mae}
  \small
  \setlength{\tabcolsep}{3.5pt}
  \begin{tabularx}{\columnwidth}{@{}r*{4}{>{\centering\arraybackslash}X}@{}}
    \toprule
    Trip & CO$_2$ MAE (g/s) & Proxy MAE (g/s) & Torque MAE (Nm) & Throttle MAE (\%)\\
    \midrule
    0  & 0.0285 & 0.0290 & 4.266 & 6.374 \\
    1  & 0.0280 & 0.0300 & 4.487 & 3.498 \\
    2  & 0.0317 & 0.0290 & 4.510 & 3.107 \\
    3  & 0.0333 & 0.0220 & 3.564 & 3.082 \\
    4  & 0.0322 & 0.0300 & 3.747 & 3.457 \\
    5  & 0.0411 & 0.0340 & 5.796 & 3.029 \\
    6  & 0.0291 & 0.0270 & 4.140 & 2.774 \\
    7  & 0.0440 & 0.0310 & 5.737 & 3.910 \\
    8  & 0.0270 & 0.0310 & 6.480 & 3.859 \\
    9  & 0.0233 & 0.0120 & 4.040 & 2.172 \\
    10 & 0.0231 & 0.0240 & 3.903 & 3.988 \\
    11 & 0.0271 & 0.0230 & 3.472 & 7.510 \\
    12 & 0.0351 & 0.0300 & 4.852 & 3.264 \\
    13 & 0.0216 & 0.0210 & 5.407 & 3.311 \\
    \bottomrule
  \end{tabularx}
\end{table}

\begin{figure*}[!t]
    \centering
    \includegraphics[width=0.49\textwidth]{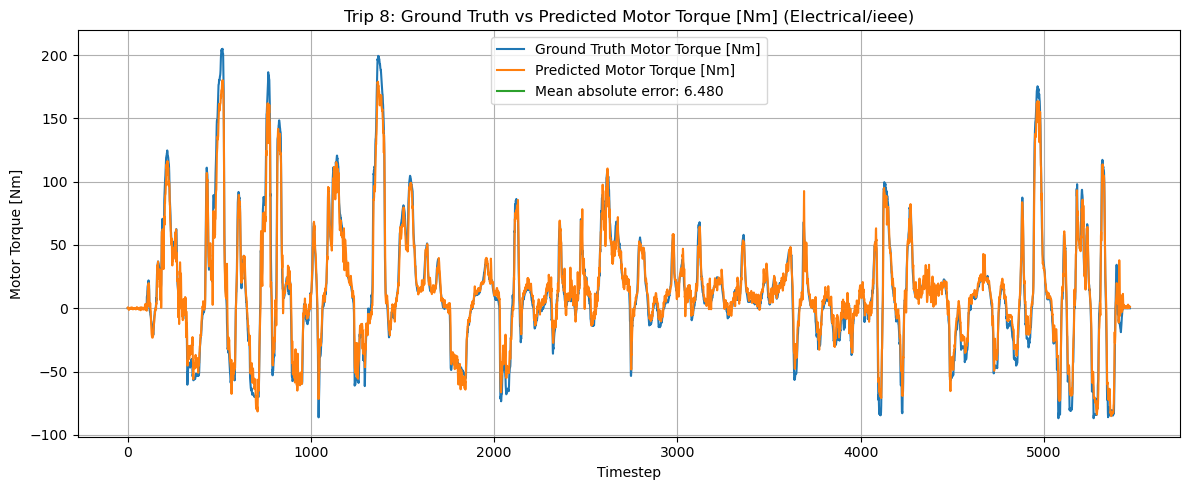}
    \includegraphics[width=0.49\textwidth]{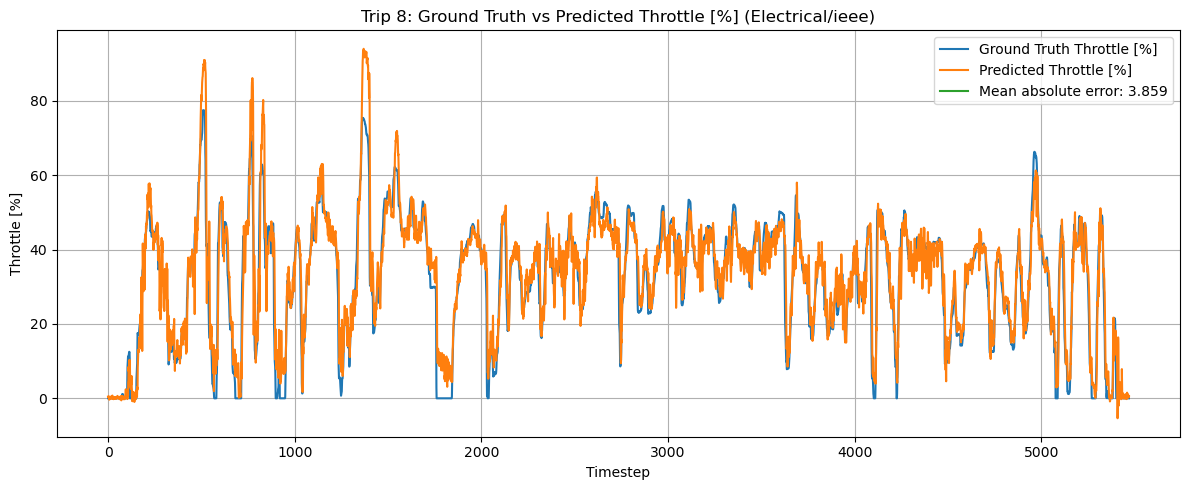}
    \caption{Held-out test (Trip 8) for the EV-Feature model: predicted vs.\ measured torque (left, Nm) and throttle (right, \%). Predictions follow ground truth closely; deviations concentrate at sharp transients.}
    \label{fig:ev_feature_test}
\end{figure*}

\begin{figure*}[!t]
    \centering
    \includegraphics[width=0.8\textwidth]{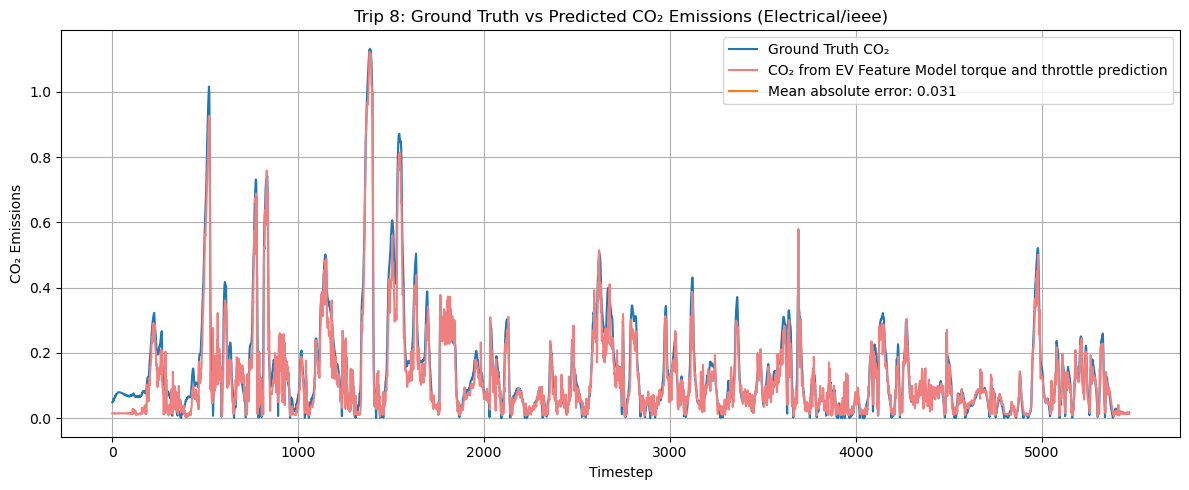}
    \caption{EV-Emissions predicted vs.\ ground-truth CO$_2$ for a representative trip (``Trip B''); mean absolute error $\approx 0.031$~g/s.}
    \label{fig:ev_emission_validation}
\end{figure*}

On held-out trips, predicted torque $(\widehat{\text{torque}})$,  and throttle $(\widehat{\text{throttle}})$ follow ground truth closely with deviations concentrated at sharp transients, while medium-scale dynamics are preserved (Figure~\ref{fig:ev_feature_test}). For a representative trip, EV-Emissions reproduces the structure of the ground-truth CO$_2$ with MAE $\approx 0.031$~g/s (Figure~\ref{fig:ev_emission_validation}). This fidelity is consistent with the aggregate training metrics and supports the use of model-predicted emissions as a faithful counterfactual under shared driving conditions.

Because EV emissions are computed via a linear map of electrical power (Eq.~\eqref{eq:e_tE}), the grid factor $\phi$ scales results linearly: a $\pm 10\%$ change in $\phi$ implies roughly $\pm 10\%$ in $e_t^{(E)}$. This clarifies how regional context (cleaner vs.\ dirtier grids) would shift absolute EV values without affecting the relative stability shown by Figure~\ref{fig:ev_training_curves} and Table~\ref{tab:ev_mae}. Moreover, our data treatment excludes negative battery current (regeneration) from the emission computation; regeneration is therefore not credited during braking, which is conservative for EV and tends to increase reported EV emissions versus net‑-energy accounting.

In general terms, these findings indicate that the EV embedded counterfactual can be constructed at test time by conditioning the EV Feature model on the observed ICEV context (speed, ambient/cabin temperatures, longitudinal acceleration) and then composing it with EV Emissions, with negligible loss of accuracy relative to using measured EV actuation—consistent with the in-domain proxy results in Table~\ref{tab:ev_mae} and Figures~\ref{fig:ev_feature_test},~\ref{fig:ev_emission_validation}. Note that proxy validation is performed on EV trips and serves as evidence of readiness for this cross-domain translation.

Although we do not report the symmetric ICEV embedded analysis, this is a data limitation rather than a methodological one. The inverse map from the shared context to engine actuation is weakly identifiable for a CVT without transmission-state/lock-up information, so we refrain from fitting it here. Still, the results we present bounded error for $f_E$ on shared inputs and negligible loss when $\hat u^{(E)}$ replaces measured actuation inside $g_E$ (Table~\ref{tab:ev_mae}, Figures~\ref{fig:ev_feature_test}–\ref{fig:ev_emission_validation})—demonstrate that, under the same preconditions on the ICEV side (e.g., explicit transmission state or a non-CVT powertrain), the same cross domain translation is feasible in the opposite direction. In that setting, a side by side plot of $\hat e^{(E)}_{1:T}$ and $e^{(C)}_{1:T}$ for a paired trajectory, together with a per-trip summary (e.g., boxplots of $\hat e^{(E)} - e^{(C)}$), presents a direct visual comparison across technologies.

%% file: conclusion.tex
This paper introduced a model neutral counterfactual framework for like for like comparison of operational CO$_2$ emissions from internal combustion (ICEV) and electric (EV) powertrains under the same observed driving context. The method keeps the measured speed profile and the shared environment fixed while learning domain specific mappings for actuation and emissions, aligning both sides on a common instantaneous metric (g/s) and isolating technology effects from confounders.

Empirically, the approach is stable and accurate. ICEV models converge without overfitting, and the EV Emissions model attains low error (MAE $\approx 0.028$~g/s). Proxy validation shows that replacing measured actuation with EV Feature predictions does not materially change accuracy (median difference $\approx -0.0025$~g/s, with proxy MAE no larger than the direct MAE in most trips). Plots at the scale of individual events confirm that the temporal structure is reproduced with small residuals, while the larger discrepancies are limited to rare periods of rapid change. Taken together, these results validate the EV embedded counterfactual as a reliable basis for assessment across technologies.

In practice, the framework supports credible comparisons using data from a single instrumented vehicle: given an ICEV trip, it produces the EV embedded emissions series for the same speed and context, enabling pointwise and trip level gap estimates without simultaneous instrumentation or repeated routes. This provides a transparent and scalable path to more robust assessments of vehicle technologies in real world driving.